%% file: ms.tex
\documentclass{article}
\usepackage{spconf,amsmath,graphicx}
\usepackage{amsmath,amssymb} 
\usepackage{epsfig}
\usepackage{multirow}
\usepackage{llncsdoc}
\usepackage{etoolbox}

\title{Combining Transfer Learning And Segmentation Information with GANs for Training Data Independent Image Registration}
\name{Dwarikanath~Mahapatra$^{1}$ and Zongyuan~Ge$^{2,3,*}$}
\address{
$^{2}$ Monash eResearch Center, Monash University, Melbourne, Australia \\
$^{3}$ AirDoc Research Australia\\
$^{1}$IBM Research Australia, Melbourne, Australia \\
$^{*}$\textit{zongyuan.ge@monash.edu} }                        

\begin{document}

\maketitle

\begin{abstract}
Registration is an important task in automated medical image analysis. Although deep learning (DL) based image registration methods out perform time consuming conventional approaches, they are heavily dependent on training data and do not generalize well for new images types. We present a DL based approach that can register an image pair which is different from the training images. This is achieved by training generative adversarial networks (GANs) in combination with segmentation information and transfer learning.  Experiments on chest Xray and brain MR images show that our method gives better registration performance over conventional methods.
\end{abstract}

 

\input{Reg_ISBI2019_Intro}

%
\input{Reg_ISBI2019_Method}

\input{Reg_ISBI2019_Expts}

%
\input{Reg_ISBI2019_Concl}

\bibliographystyle{IEEEbib}
\bibliography{Reg_ISBI2019_Ref}

\end{document}

%% file: Reg_ISBI2019_Intro.tex

\section{Introduction}
\label{sec:intro}

Important medical image analysis tasks such as  atlas building, and monitoring pathological changes over multiple patient visits have deformable image registration as an essential step. Iterative gradient descent methods used in conventional registration methods are slow in practice while deep learning (DL) methods can be very fast at test time. Most DL based methods rely on large datasets for training. Since it is difficult to obtain ground truth data for registration it restricts the method's efficacy on new image types and in real world scenarios.
A network trained to register a pair of chest xray images does not  perform equally well on a pair of brain magnetic resonance (MR) images, or Xray images from other scanners. Although conventional registration methods are time consuming, their performance is consistent across different image types. Thus DL based methods have to be retrained for novel images.  In this paper we address this challenge by proposing a DL based method that, once trained on a particular dataset, can be easily used on other image pairs without extensive retraining.



A comprehensive review of conventional medical image registration methods can be found in \cite{RegRev}. Previous approaches to DL based image registration involve the use of convolutional stacked autoencoders (CAE) \cite{WuTBME}, convolutional neural network (CNN) regressors \cite{Miao_Reg,FlowNet,Vos_DIR}, and  CNNs with reinforcement learning \cite{Liao_Reg}. These approaches use a conventional model to generate the transformed image from the predicted deformation field which increases computation time and does not fully utilize the generative capabilities of DL methods. 
CNNs trained on simulated deformations were used in \cite{RegNet} while in \cite{BalaCVPR18} a parameterized registration function is learned from training data, and does not require ground truth.

The above methods are limited by the need of spatially corresponding patches or being too dependent on training data. Generative models can overcome some of these limitations by generating the registered image and the deformation field. In previous work \cite{MahapatraGANISBI2018} we used generative adversarial networks (GANs) for multimodal retinal image registration, and in \cite{Mahapatra_MLMI18} show the advantages of including segmentation for registration compared to conventional registration. 
In this paper we build on our previous works and show how segmentation information can be leveraged to design a DL registration method that does not require extensive retraining when used with different datasets. Our primary contribution is in using principles of transfer learning for achieving dataset independent registration. We show that our method, despite being trained on chest Xray images, achieves high performance levels with test images of brain MRI.
  %

%

%% file: Reg_ISBI2019_Method.tex

\section{Methods}
\label{sec:met}

In our proposed method the generator network, $G$, takes two input images: 1) reference image ($I^{Ref}$), and 2) floating image ($I^{Flt}$) to be registered to $I^{Ref}$. 
The outputs of $G$ are: $I^{Trans}$, the registered image (transformed version of $I^{Flt}$); 2) $I^{Def-Recv}$ the recovered deformation field and 3) $I^{Trans}_{Seg}$ the segmentation mask of $I^{Trans}$.
%
Our method has two parts: 1) model training using segmentation information. We call this part segmented augmented registration (SAR); 2) model finetuning for a new test image pair using transfer learning. 

\subsection{Segmentation Augmented Registration Using GANs}

GANs \cite{GANs} are generative models where the generator $G$ outputs a desired image type while a discriminator $D$ outputs a probability of the generated image matching the training data.
The training database has chest Xray images and the corresponding masks of  two lungs. To generate training data the floating images are first affinely aligned to the respective  reference images. 
The aligned images are subjected to local elastic deformation using B-splines with the pixel displacements in the range of $\pm[1,20]$. We denote this deformation field as $I_{Def-App}$, the applied deformation field. 
%
The original images are $I^{Ref}$ and the transformed images are $I^{Flt}$.
$G$'s parameters $\theta_G$ are given by, 
\begin{equation}
\widehat{\theta}=\arg \min_{\theta_G} \frac{1}{N} \sum_{n=1}^{N} l^{SAR}\left(G_{\theta_G}(I^{Flt}),I^{Ref},I^{Flt},I^{Ref}_{Seg}\right),
\label{eq:theta1}
\end{equation}
where the loss function $l^{SAR}$ combines content loss (Eqn.~\ref{eq:conLoss}) and adversarial loss (Eqn.~\ref{eqn:cyGan1}), and $G_{\theta_G}(I^{Flt})=I^{Trans}$. The content loss is given by 
\begin{equation}
\begin{split}
& l_{content}(I^{Trans},I^{Ref}) = NMI (I^{Ref},I^{Trans}) + \\
& \left[1-SSIM(I^{Ref},I^{Trans})\right] + VGG(I^{Ref},I^{Trans}). 
\end{split}
\label{eq:conLoss}
\end{equation} 
%

$NMI$ denotes normalized mutual information between $I^{Ref}$ and $I^{Trans}$ and $SSIM$ denotes structural similarity index metric (SSIM). 
$VGG$ is the $L2$ distance between two images using all the multiple feature maps obtained from a pre-trained $VGG16$ network \cite{VGG}. 
This sums up to $64\times2+128\times2+256\times2+512\times3+512\times3=3968$ feature maps and compares information from multiple scales for better robustness. All feature maps are normalized to values between $[0,1]$. 

\begin{figure}[t]
\begin{tabular}{c}
\includegraphics[height=3.9cm, width=7.99cm]{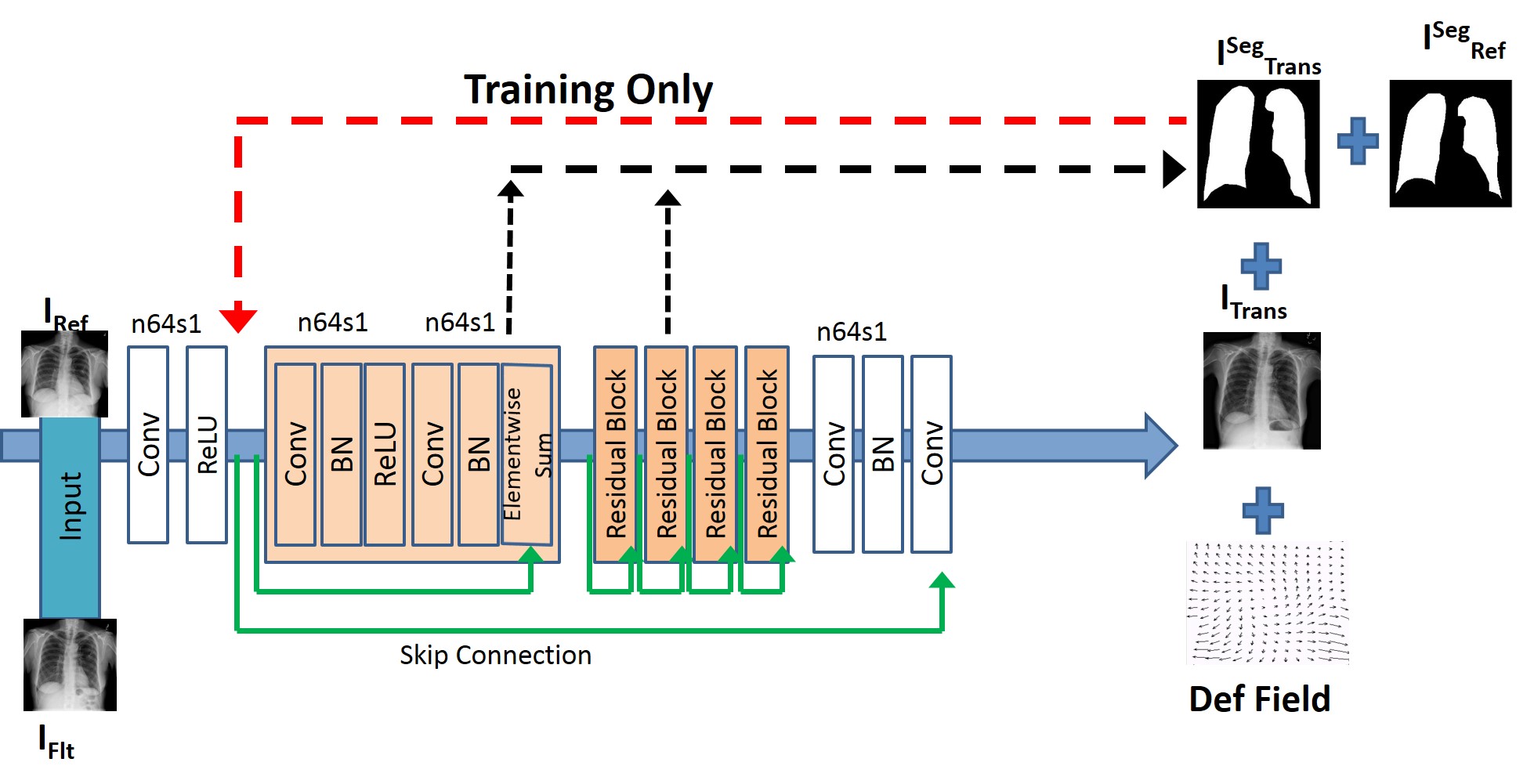}  \\ 
(a) \\
\includegraphics[height=3.9cm, width=7.99cm]{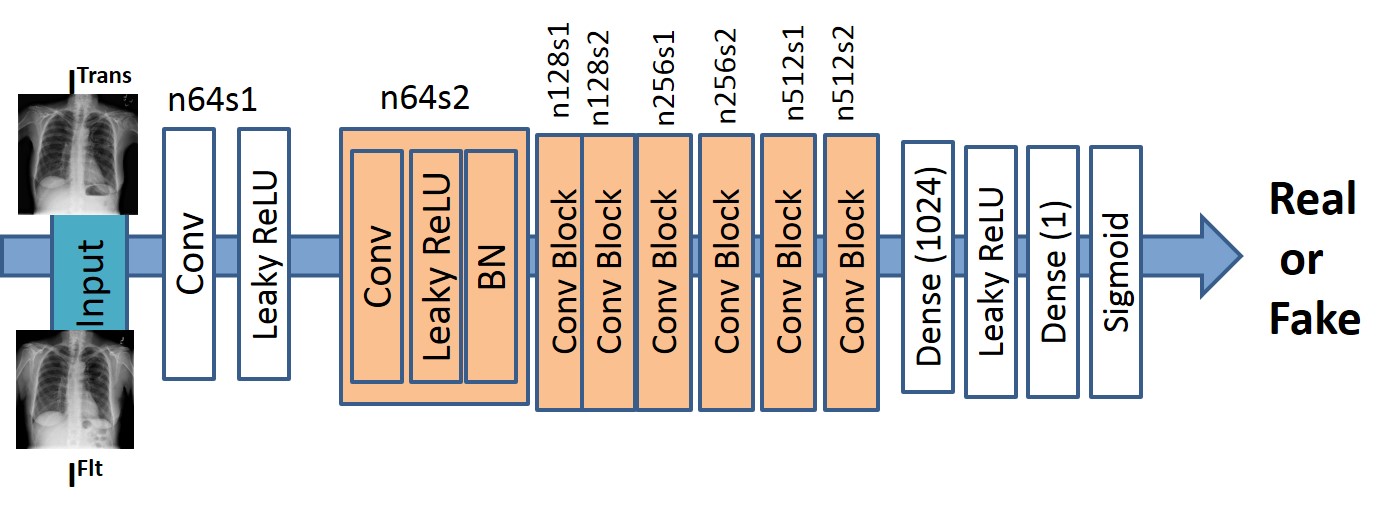}  \\
(b)\\
\end{tabular}
\caption{(a) Generator Network; (b) Discriminator network. $n64s1$ denotes $64$ feature maps (n) and stride (s) $1$ for each convolutional layer.}
\label{fig:Gan}
\end{figure}

\begin{figure}[t]
\begin{tabular}{ccc}
 \includegraphics[height=2.5cm, width=2.5cm]{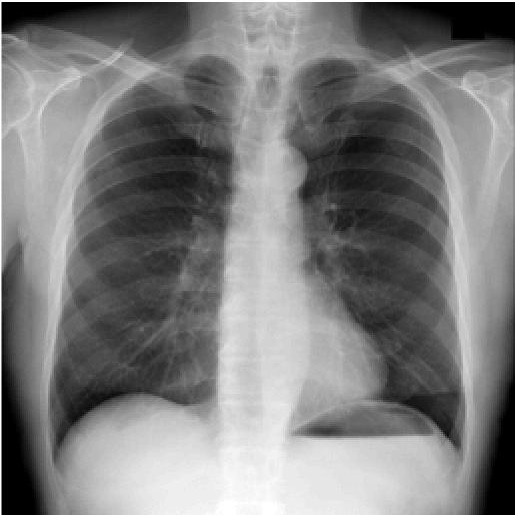} & 
\includegraphics[height=2.5cm, width=2.5cm]{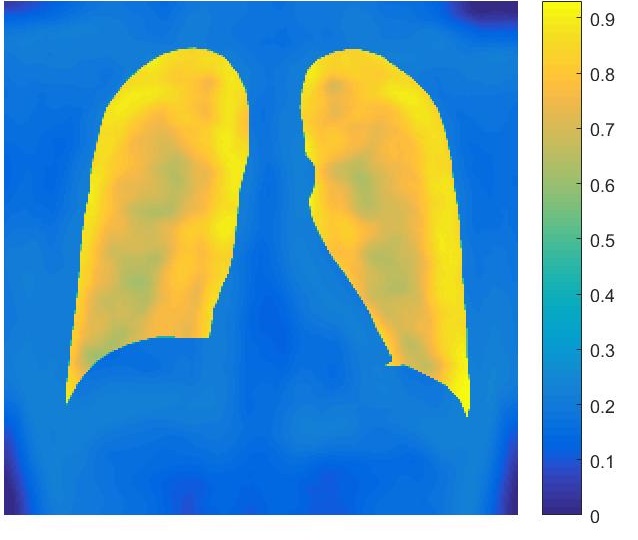}  &
\includegraphics[height=2.5cm, width=2.5cm]{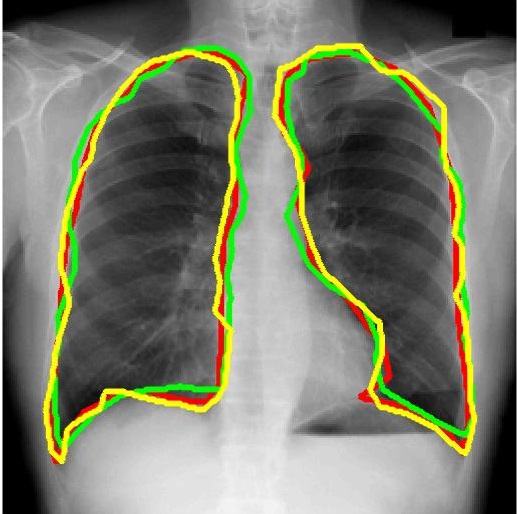}  \\ 
(a) & (b) & (c) \\
\end{tabular}
\caption{Example segmentation output from generator network on the fly. (a) original image; (b) fused weighted convolutional map; (c) segmentation outputs - green contour shows mask obtained by Otsu's thresholding the image in (b), red contour is the manual segmentation, yellow contour is the output of UNet segmentation.   }
\label{fig:genmask}
\end{figure}

Figure~\ref{fig:Gan}(a) shows the generator network $G$ which employs residual blocks, each block having two convolutional layers with $3\times3$ filters and $64$ feature maps, followed by batch normalization and ReLU activation. $G$ also outputs a deformation field and the segmentation masks of $I^{Trans},I^{Ref}$. The segmentation masks are obtained by fusing the weighted normalized output maps of the different convolution layers and applying Otsu's thresholding. The convolution layer outputs highlight the different anatomies of the image and the weights quantify the importance of each map. Figure~\ref{fig:genmask} shows an example image, the obtained fused convolution mask and the segmented mask by Otsu's thresholding (along with a contour of the manual segmentation and the output of UNet segmentation). This shows that our segmentation mask is very similar to UNet's output.

The discriminator $D$ (Figure~\ref{fig:Gan} (b)) has eight convolutional layers with the kernels increasing by a factor of $2$ from $64$ to $512$ . Leaky ReLU is used and strided convolutions reduce the image dimension  when the number of features is doubled. The resulting $512$ feature maps are
followed by two dense layers and a final sigmoid activation to obtain a probability map. $D$ evaluates similarity of intensity distribution between $I^{Trans}$ and $I^{Ref}$, the accuracy of the two segmentation masks compared to their manual counterparts and the  overlap between $I^{Trans}_{Seg}$ and $I^{Ref}_{Seg}$, and the error between generated and reference deformation fields.


\subsection{Adversarial Loss with Segmentation Information}

In addition to the content loss (Eqn~\ref{eq:conLoss}) we have: 1) an adversarial loss; 
and 2) a cycle consistency loss to ensure transformations $G,F$ do not contradict each other.
Since our generator network has multiple outputs we have additional terms for the adversarial loss. The first term matches the distribution of $I^{Trans}$ to  $I^{Flt}$  
and is given by:
\begin{equation}
\begin{split}
 L_{cycGAN}(G,D_Y) & = E_{y\in p_{data}(y)} \left[\log D_Y(y)\right] + \\
& E_{x\in p_{data}(x)} \left[\log \left(1-D_Y(G(x))\right)\right], \\ 
\end{split}
\label{eqn:cyGan1}
\end{equation}
where $X=I^{Flt}$ and $Y=I^{Ref}$.
%
 $L_{cycGAN}(F,D_X)$ is the corresponding adversarial loss for $F$ and $D_X$. 
%
%
The cycle consistency loss \cite{CyclicGANs} ensures the deformation fields are reversible and is achieved by, 
%
%
\begin{equation}
L_{cyc}(G,F)= E_{x} \left\|F(G(x))-x\right\|_1 + E_{y} \left\|G(F(y))-y\right\|_1,
\label{eqn:cyGan2}
\end{equation}
%


Segmentation information is included in the adversarial loss by calculating the logarithm of the dice metric (DM) between the generated mask $I^{Trans}_{Seg}$ during each training step and $I^{Ref}_{Seg}$ the segmentation mask  of $I^{Ref}$. 
$I^{Trans}_{Seg}$ is obtained by applying $I^{Def-Recv}$, the  recovered deformation field, to $I^{Flt}_{Seg}$. 
The third adversarial loss term is the mean square error between $I^{Def-App}$ and $I^{Def-Recv}$, the applied and recovered deformation fields. 
The final adversarial loss is 
\begin{equation}
\begin{split}
L_{adv}= L_{cycGAN}(G,D_{I^{Ref}}) + L_{cycGAN}(F,D_{I^{Flt}}) + & \\  
\log DM(I^{Ref}_{Seg},I^{Trans}_{Seg}) + & \\
\log \left(1-MSE_{Norm}(I^{Def-App},I^{Def-Recv}) \right), & 
\end{split}
\label{eqn:adloss}
\end{equation}
where $MSE_{Norm}$ is the MSE normalized to $[0, 1]$, and $1-MSE_{Norm}$ ensures that similar deformation fields gives a corresponding higher value. All terms in the adversarial loss function have values in $[0,1]$. 
 The full objective function is
\begin{equation}
L(G,F,D_{I^{Flt}},D_{I^{Ref}})= L_{adv} + l_{content} + \lambda L_{cyc}(G,F) 
\label{eqn:cyGan3}
\end{equation}
where $\lambda=10$ controls the contribution of the two objectives. The optimal parameters are given by:
\begin{equation}
G^{*},F^{*}=\arg \min_{F,G} \max_{D_{I^{Flt}},D_{I^{Ref}}} L(G,F,D_{I^{Flt}},D_{I^{Ref}})
\label{eqn:CyGan4}
\end{equation}

\subsection{Registering a new image}
\label{met:newimg}

The primary advantage of our method is its ability to register a new image pair. During training the trained network $G$ generates the segmentation masks of the input images, the registered image and the deformation field. Since we do not have applied deformation field to determine the accuracy of test image, the best way to gauge registration performance is by comparing $I^{Trans}$ and $I^{Ref}$  based on feature maps and segmentation mask output.  If the input test image pair consists of lung images (with training images also of the lung) the generated outputs will be close to the desired values and there is no need for any finetuning of the weights.

The challenge lies in registering a completely new image pair, e.g., brain MR images which we achieve by transfer learning. In transfer learning for image classification the weights of all except the last few layers are frozen. We use a similar principle to register a new test image pair. The weights of the last convolution layer of generator $G$ are updated iteratively based on the output of the discriminator. In the case of registering a new test image pair, the network is being fine-tuned. As a result, the discriminator network also comes into play. However, in this case the adversarial loss is based on the cyclic GAN and segmentation mask loss terms, i.e., the first three terms of Eqn.~\ref{eqn:adloss}, and the deformation field loss term is excluded. The weight updates occur till the difference of cost function values for consecutive iterations is less than $1\%$.  The update happens for $10-30$ iterations depending on the input image pair. Since the weight updates are only for the last layer and the computation is GPU based, the time taken for registration is very low, around $0.3-0.5$ seconds.

%% file: Reg_ISBI2019_Expts.tex

\section{Experiments}
\label{sec:expts}


Our registration method was trained on the NIH ChestXray14 dataset \cite{NIHXray}. The original dataset contains $112,120$ frontal-view X-rays with $14$ disease labels. To make it suitable for registration we selected images from  $50$ with multiple visits. The first visit image was $I^{Ref}$ and subsequent images were $I^{Flt}$. In total we selected $906$ images where each patient had minimum  $3$ images and maximum $8$ images. The left and right lung were manually outlined in each image. All images were resized to $512\times512$ pixels before the manual annotations. 
For all our experiments we split the dataset into training, validation and test sets comprising of $70,10,20 \%$ of he images. The split was done at the patient level such that images from a single patient were in one fold only. All the reported results are for the test set. 
%
%
Registration performance was validated using mean absolute distance (MAD), the $95\%$ Hausdorff Distance ($HD_{95}$) and Dice Metric (DM). 
%
After  training on chest xray images we apply our method to brain and cardiac MR images with finetuning.


Our method was implemented in TensorFlow using Adam \cite{Adam} with $\beta_1=0.93$ and batch normalization. The generator network $G$ was trained with a learning rate of $0.001$ and $10^{5}$ update iterations. Mean square error (MSE) based ResNet was used to initialize $G$. The final GAN was trained with $10^{5}$ update iterations at learning rate $10^{-3}$. Training and test was performed on a NVIDIA Tesla K$40$ GPU with $12$ GB RAM. 

We show results for: 1) $SARNet$ - our proposed registration network; 2) $SAR_{NoSeg}$ - $SAR$ without using segmentation information; 3) $FlowNet$ - the registration method of \cite{FlowNet}; 4) $DIRNet$ - the method of \cite{Vos_DIR}; 5) $Voxel~Morph$ - the registration method of \cite{BalaCVPR18};  and 6) a conventional registration method Elastix \cite{Elastix}. All networks were trained on lung images and applied to brain images.
%
The average training time for an augmented dataset with $98,000$ images is $36$ hours. 
The following parameter settings were used for Elastix: non rigid registration using normalized mutual information (NMI) as the cost function. Nonrigid transformations are modelled by B-splines \cite{FFD}, embedded in a multi-grid setting. The grid spacing was set to $ 80,40,20,10,5$ mm with the corresponding downsampling factors being $4, 3, 2, 1, 1$.

\subsection{Registration Results For Brain MRI}
\label{expt:atlas}

We use the $800$ images of the ADNI-1 dataset \cite{Bala33} consisting of $200$ controls, $400$ MCI and $200$ Alzheimer's Disease patients. The MRI protocol for ADNI1 focused on consistent longitudinal structural imaging on $1.5T$ scanners using $T1$ and dual echo $T2-$weighted sequences.  
All scans were resampled to $256\times256\times256$ with $1$mm isotropic voxels. Pre-processing includes affine registration and brain extraction using FreeSurfer \cite{Bala17}. 
%
%
%
The atlas is an average of multiple volumes and obtained by aligning MR volumes from \cite{Bala17}. 

\begin{figure}[t]
\begin{tabular}{ccc}
\includegraphics[height=2.5cm,width=2.5cm]{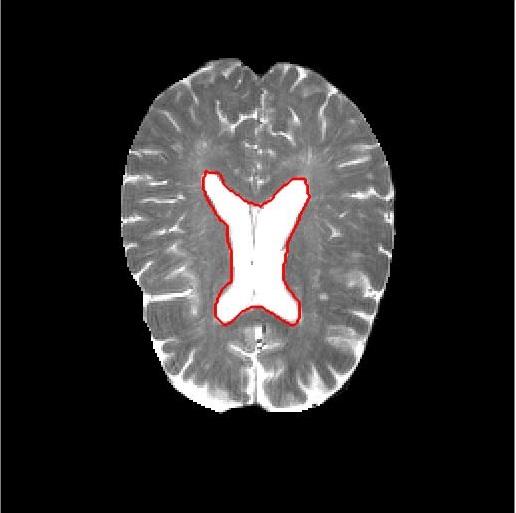} &
\includegraphics[height=2.5cm,width=2.5cm]{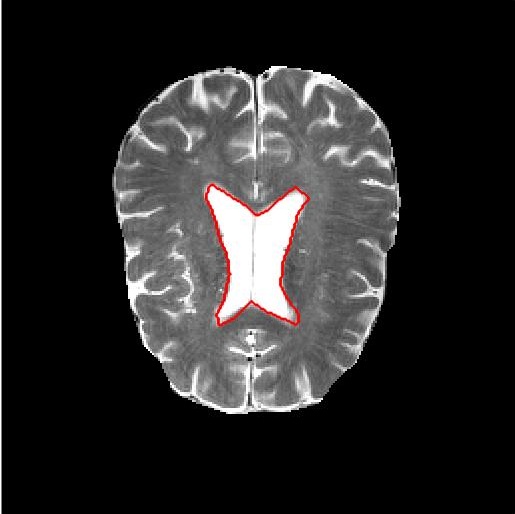} &
\includegraphics[height=2.5cm,width=2.5cm]{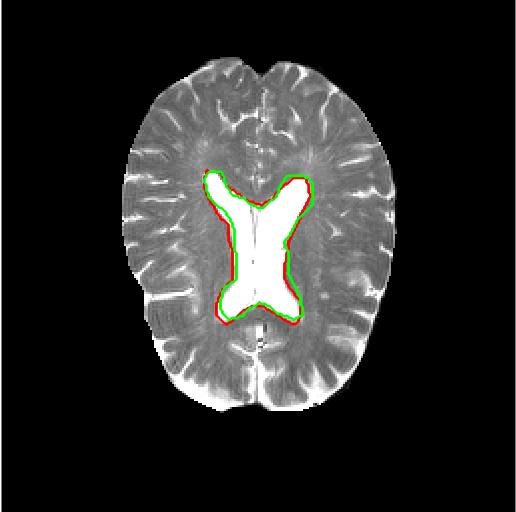} \\
(a) & (b) & (c)  \\
\includegraphics[height=2.5cm,width=2.5cm]{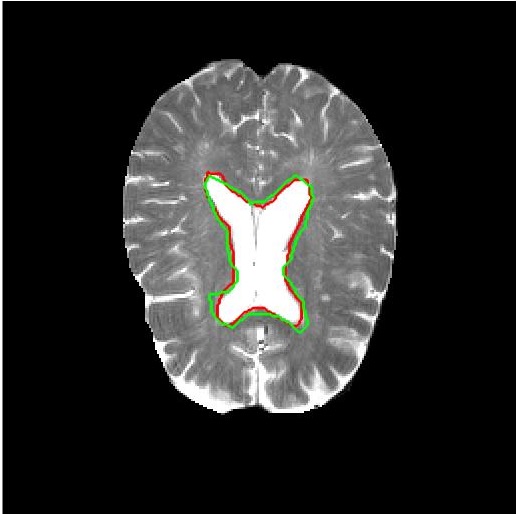} &
\includegraphics[height=2.5cm,width=2.5cm]{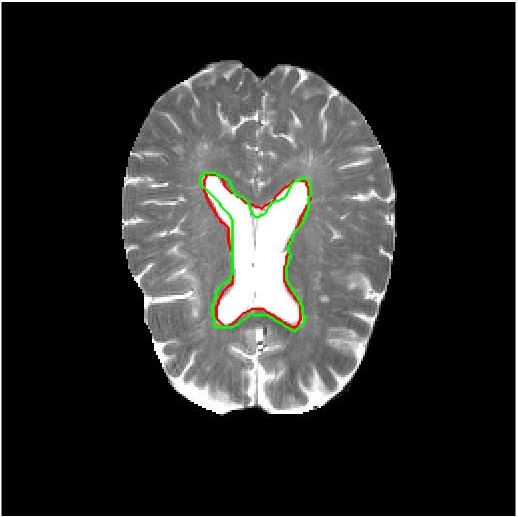} &
\includegraphics[height=2.5cm,width=2.5cm]{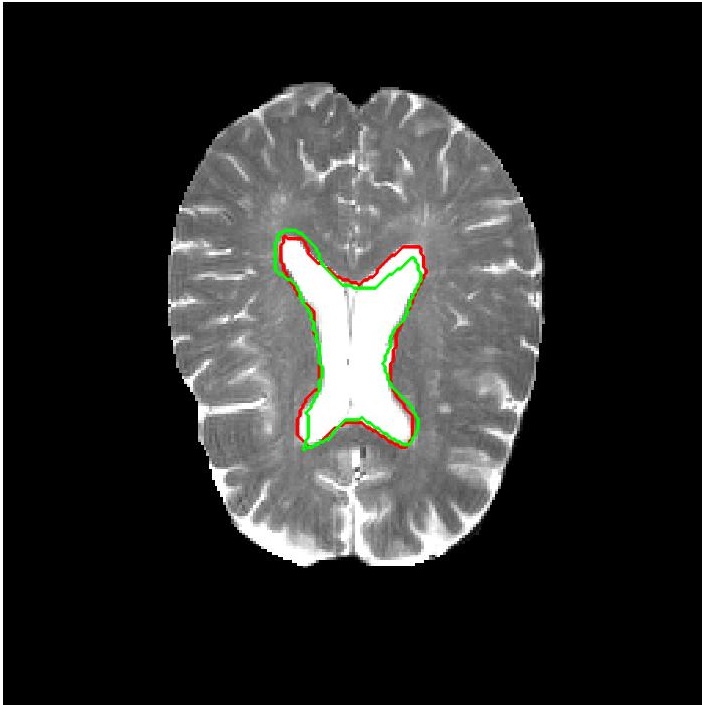} \\
(d) & (e) & (f) \\ 
\end{tabular}
\caption{Results for atlas based brain MRI image registration. (a) $I_{Ref}$ with $I_{Ref}^{Seg}$ (b) $I_{Flt}$ with $I_{Flt}^{Seg}$. Superimposed registered mask (in green) obtained using: (c) $SAR-Net$; (d)\cite{BalaCVPR18}; (e) $SAR_{NoSeg}$; and (f) $DIR-Net$.}
\label{fig:brain}
\end{figure}

We show the reference image (or the atlas image) in  Figure~\ref{fig:brain} (a) followed by an example floating image in Figure~\ref{fig:brain} (b). The ventricle structure to be aligned is shown in red in both images. Figures~\ref{fig:brain} (c)-(i) show the deformed  structures obtained by applying the registration field obtained from different methods to the floating image and superimposing these structures on the atlas image. The deformed structures from the floating image are shown in green. In case of a perfect registration the green and red contours should coincide. 

Table~\ref{tab:brain} shows the results of brain registration before and after registration. Our method and \cite{BalaCVPR18} perform the best, with ours better. The results clearly demonstrate that our method can effectively transfer learned information from one dataset to another. All other methods have been trained on the brain images. Despite that fact our method outperforms them indicating the importance of using segmentation information in better registration. When the training and test images are of different types then the final registration output requires between $10-30$ iterations but the time is not noticeable since our experiments are done on GPUs.


\begin{table}[t]
\begin{tabular}{|c|c|c|c|c|}
\hline
{} & {Bef.} & \multicolumn {3}{|c|}{After Registration}  \\ \cline{3-5} 
{} & {Reg} & {SAR}  & {SAR} & {DIR}   \\ 
{} & {} & {Net}  & {$_{NoSeg}$} & {Net}   \\ \hline
%
{DM($\%$)} & {67.2} & {74.1}  & {71.2} & {70.9} \\ \hline
{HD$_{95}$(mm)} & {14.5} & {10.8}  & {12.3} & {12.7} \\ \hline
{MAD} & {16.1} & {12.0}  & {13.6} & {14.1} \\ \hline
{Time(s)} & {} & {0.5} & {0.4} & {0.6} \\ \hline
{} & \multicolumn {4}{|c|}{After Registration}  \\ \cline{2-5} 
{} & {Flow} & {GC} & {Elastix} & {Voxel} \\ 
{} & {Net} & {-SAR} & {} & {Morph} \\  \hline
%
{DM($\%$)} & {69.4} & {70.2} & {70.3} & {71.3} \\ \hline
{HD$_{95}$(mm)} & {14.1} & {13.8} & {14.2} & {13.6} \\ \hline
{MAD} & {15.1} & {15.3} & {15.9} & {14.2} \\ \hline
{Time(s)} & {0.5} & {0.6} & {21} & {0.5} \\ \hline
\end{tabular}
\caption{Registration results for brain images when network is trained on lung images. }
\label{tab:brain}
\end{table}

\textbf{Baseline Performance}
Table~\ref{tab:newtest2} summarizes performance using brain MRI for training and test. A 5-fold cross validation uses all the $800$ images as part of the test set exactly once which provides a fair comparison with the numbers in Table~\ref{tab:brain}. Results show that transfer learning gives results similar to when the training and test images are of the same type. This proves the efficacy of our proposed transfer learning based image registration.

\begin{table}[t]
\begin{tabular}{|c|c|c|c|c|}
\hline
{} & {Bef.} & \multicolumn {3}{|c|}{After Registration}  \\ \cline{3-5} 
{} & {Reg} & {SAR}  & {SAR} & {DIR}   \\ 
{} & {} & {Net}  & {$_{NoSeg}$} & {Net}   \\ \hline
%
{DM($\%$)} & {67.2} & {74.9}  & {72.0} & {71.7} \\ \hline
{HD$_{95}$(mm)} & {14.5} & {10.3}  & {11.9} & {12.1} \\ \hline
{MAD} & {16.1} & {11.4}  & {13.1} & {13.7} \\ \hline
{Time(s)} & {} & {0.5} & {0.4} & {0.6} \\ \hline
{} & \multicolumn {4}{|c|}{After Registration}  \\ \cline{2-5} 
{} & {Flow} & {GC} & {Elastix} & {Voxel} \\ 
{} & {Net} & {-SAR} & {} & {Morph} \\  \hline
%
{DM($\%$)} & {70.6} & {70.9} & {71.2} & {72.1} \\ \hline
{HD$_{95}$(mm)} & {13.8} & {13.4} & {13.5} & {13.0} \\ \hline
{MAD} & {14.6} & {14.8} & {15.4} & {13.8} \\ \hline
{Time(s)} & {0.5} & {0.6} & {21} & {0.5} \\ \hline
\end{tabular}
\caption{Registration results for brain images when network is trained on brain images. }
\label{tab:newtest2}
\end{table}


%% file: Reg_ISBI2019_Concl.tex

\section{Conclusion}
\label{sec:concl}

We propose a novel deep learning framework to register brain MR images by training on lung Xray images. We leverage segmentation information and transfer learning with generative adversarial networks. Experimental results show our approach achieves registration with almost similar accuracy as one would obtain when the training and test dataset consist of similar images. 


%% file: ms.bbl
\begin{thebibliography}{10}

\bibitem{RegRev}
J.B.A. maintz and M.A. Viergever,
\newblock ``A survey of medical image registration,''
\newblock {\em Med. Imag. Anal}, vol. 2, no. 1, pp. 1--36, 1998.

\bibitem{WuTBME}
G.~Wu, M.~Kim, Q.~Wang, B.~C. Munsell, , and D.~Shen.,
\newblock ``Scalable high performance image registration framework by
  unsupervised deep feature representations learning.,''
\newblock {\em IEEE Trans. Biomed. Engg.}, vol. 63, no. 7, pp. 1505--1516,
  2016.

\bibitem{Miao_Reg}
S.~Miao, Y.~Zheng Z.J.~Wang, and R.~Liao,
\newblock ``Real-time 2d/3d registration via cnn regression,''
\newblock in {\em IEEE ISBI}, 2016, pp. 1430--1434.

\bibitem{FlowNet}
A.~Dosovitskiy, P.~Fischer, and et. al.,
\newblock ``Flownet: Learning optical flow with convolutional networks,''
\newblock in {\em In Proc. IEEE ICCV}, 2015, pp. 2758--2766.

\bibitem{Vos_DIR}
B.~de~Vos, F.~Berendsen, M.A. Viergever, M.~Staring, and I.~Isgum,
\newblock ``End-to-end unsupervised deformable image registration with a
  convolutional neural network,''
\newblock in {\em arXiv preprint arXiv:1704.06065}, 2017.

\bibitem{Liao_Reg}
R.~Liao, S.~Miao, P.~de~Tournemire, S.~Grbic, A.~Kamen, T.~Mansi, and
  D.~Comaniciu,
\newblock ``An artificial agent for robust image registration,''
\newblock in {\em AAAI}, 2017, pp. 4168--4175.

\bibitem{RegNet}
H.~Sokooti, B.~de~Vos, F.~Berendsen, B.P.F. Lelieveldt, I.~Isgum, and
  M.~Staring,
\newblock ``Nonrigid image registration using multiscale 3d convolutional
  neural networks,''
\newblock in {\em MICCAI}, 2017, pp. 232--239.

\bibitem{BalaCVPR18}
G.~Balakrishnan, A.~Zhao, M.R. Sabuncu, and J.~Guttag,
\newblock ``An supervised learning model for deformable medical image
  registration,''
\newblock in {\em Proc. CVPR}, 2018, pp. 9252--9260.

\bibitem{MahapatraGANISBI2018}
D.~Mahapatra, B.~Antony, S.~Sedai, and R.~Garnavi,
\newblock ``Deformable medical image registration using generative adversarial
  networks,''
\newblock in {\em In Proc. IEEE ISBI}, 2018, pp. 1449--1453.

\bibitem{Mahapatra_MLMI18}
D.~Mahapatra, Z.~Ge, S.~Sedai, and R.~Chakravorty.,
\newblock ``Joint registration and segmentation of xray images using generative
  adversarial networks,''
\newblock in {\em In Proc. MICCAI-MLMI}, 2018, pp. 73--80.

\bibitem{GANs}
I.~Goodfellow, J.~Pouget-Abadie, M.~Mirza, B.~Xu, D.~Warde-Farley, S.~Ozair,
  A.~Courville, and Y.~Bengio,
\newblock ``Generative adversarial nets,''
\newblock in {\em Proc. NIPS}, 2014, pp. 2672--2680.

\bibitem{VGG}
K.~Simonyan and A.~Zisserman.,
\newblock ``Very deep convolutional networks for large-scale image
  recognition,''
\newblock {\em CoRR}, vol. abs/1409.1556, 2014.

\bibitem{CyclicGANs}
J.Y. Zhu, T.park, P.~Isola, and A.A. Efros,
\newblock ``Unpaired image-to-image translation using cycle-consistent
  adversarial networks,''
\newblock in {\em arXiv preprint arXiv:1703.10593}, 2017.

\bibitem{NIHXray}
X.~Wang, Y.~Peng, L.~Lu, Z.~Lu, M.~Bagheri, and R.M. Summers,
\newblock ``Chestx-ray8: Hospital-scale chest x-ray database and benchmarks on
  weakly-supervised classification and localization of common thorax
  diseases,''
\newblock in {\em In Proc. CVPR}, 2017.

\bibitem{Adam}
D.P. Kingma and J.~Ba,
\newblock ``Adam: A method for stochastic optimization,''
\newblock in {\em arXiv preprint arXiv:1412.6980,}, 2014.

\bibitem{Elastix}
S.~Klein, M.~Staring, K.~Murphy, M.A. Viergever, and J.P.W. Pluim.,
\newblock ``Elastix: a toolbox for intensity based medical image
  registration.,''
\newblock {\em IEEE Trans. Med. Imag..}, vol. 29, no. 1, pp. 196--205, 2010.

\bibitem{FFD}
D.~Rueckert, L.I Sonoda, C.~Hayes, D.L.G Hill, M.O Leach, and D.J Hawkes.,
\newblock ``Nonrigid registration using free-form deformations: application to
  breast mr images.,''
\newblock {\em IEEE Trans. Med. Imag..}, vol. 18, no. 8, pp. 712--721, 1999.

\bibitem{Bala33}
S.~G. Mueller,
\newblock ``Ways toward an early diagnosis in alzheimer’s disease: the
  alzheimer’s disease neuroimaging initiative ({ADNI}).,''
\newblock {\em Alzheimer’s \& Dementia}, vol. 1, no. 1, pp. 55--66, 2005.

\bibitem{Bala17}
B.~Fischl,
\newblock ``Freesurfer,''
\newblock {\em Neuroimage}, vol. 62, no. 2, pp. 774--781, 2015.

\end{thebibliography}
